%% file: main.tex
\documentclass[conference]{IEEEtran}
\IEEEoverridecommandlockouts
\usepackage{cite}
\usepackage{amsmath,amssymb,amsfonts}
\usepackage{algorithmic}
\usepackage{graphicx}
\usepackage{textcomp}
\usepackage{xcolor}
\usepackage{booktabs}
\usepackage{multirow}
\usepackage{hyperref}
\def\BibTeX{{\rm B\kern-.05em{\sc i\kern-.025em b}\kern-.08em
    T\kern-.1667em\lower.7ex\hbox{E}\kern-.125emX}}
\begin{document}

\title{MHDash: An Online Platform for Benchmarking Mental Health–Aware AI Assistants

% {\footnotesize \textsuperscript{*}Note: Sub-titles are not captured in Xplore and should not be used}
}

% \author{Anonymous Author(s)}

\author{\IEEEauthorblockN{Yihe Zhang\thanks{This is the author version of an accepted paper. 
The final version will appear in IEEE SoutheastCon 2026.}}
\IEEEauthorblockA{\textit{Informatics Research Institute}\\
\textit{Univeristy of Louisiana at Lafayette} \\
yihe.zhang@louisiana.edu}
\and
\IEEEauthorblockN{Cheyenne N Mohawk}
\IEEEauthorblockA{\textit{Department of Psychology} \\
\textit{Univeristy of Louisiana at Lafayette} \\
cheyenne.mohawk1@louisiana.edu}
\and
\IEEEauthorblockN{Kaiying Han}
\IEEEauthorblockA{\textit{Informatics Research Institute}\\
\textit{Univeristy of Louisiana at Lafayette} \\
kaiying.han1@louisiana.edu}
\and
\IEEEauthorblockN{Vijay Srinivas Tida}
\IEEEauthorblockA{\textit{Department of Computer Science} \\
\textit{CSB/SJU} \\
vtida001@csbsju.edu}
\and
\IEEEauthorblockN{Manyu Li}
\IEEEauthorblockA{\textit{Department of Psychology}\\
\textit{Univeristy of Louisiana at Lafayette} \\
manyu.li@louisiana.edu}
\and
\IEEEauthorblockN{Xiali Hei}
\IEEEauthorblockA{\textit{School of Computing and Informatics} \\
\textit{Univeristy of Louisiana at Lafayette}\\
xiali.hei@louisiana.edu}
}

% \and
% \and
% \and
% \and
% \and

\maketitle

\begin{abstract}
Large language models (LLMs) are increasingly applied in mental health support systems, where reliable recognition of high-risk states such as suicidal ideation and self-harm is safety-critical. However, existing evaluations primarily rely on aggregate performance metrics, which often obscure risk-specific failure modes and provide limited insight into model behavior in realistic, multi-turn interactions.
We present MHDash, an open-source platform designed to support the development, evaluation, and auditing of AI systems for mental health applications. MHDash integrates data collection, structured annotation, multi-turn dialogue generation, and baseline evaluation into a unified pipeline. The platform supports annotations across multiple dimensions, including Concern Type, Risk Level, and Dialogue Intent, enabling fine-grained and risk-aware analysis.
Our results reveal several key findings: (i) simple baselines and advanced LLM APIs exhibit comparable overall accuracy yet diverge significantly on high-risk cases; (ii) some LLMs maintain consistent ordinal severity ranking while failing absolute risk classification, whereas others achieve reasonable aggregate scores but suffer from high false negative rates on severe categories; and (iii) performance gaps are amplified in multi-turn dialogues, where risk signals emerge gradually. These observations demonstrate that conventional benchmarks are insufficient for safety-critical mental health settings.
By releasing MHDash as an open platform, we aim to promote reproducible research, transparent evaluation, and safety-aligned development of AI systems for mental health support.\footnotemark
\end{abstract}

\footnotetext{MHDash is publicly available at \url{https://mhdash.socialshields.org/}.}


\input{1-introduction}
\input{2-related}
\input{3-system}
\input{4-data}

% \input{5-benchmark}
\input{6-evaluation}
\input{7-conclusion}

\bibliographystyle{IEEEtran}
\bibliography{main}

\end{document}

%% file: 1-introduction.tex
\section{Introduction}
Large language models (LLMs) are increasingly deployed as conversational agents in mental health support scenarios, including emotional counseling, crisis intervention, and peer support platforms. Their ability to generate empathetic, context-aware responses has led to rapid adoption across online forums, chatbots, and digital health applications. However, this growing reliance on LLM-driven systems introduces significant safety and security risks, particularly when models fail to correctly identify or respond to high-risk mental health states such as suicidal ideation or self-harm behavior.

Unlike conventional NLP tasks, mental health support operates in a safety-critical setting where misclassification can have severe real-world consequences. False negative cases in which high-risk users are incorrectly assessed as low risk may delay or prevent timely intervention. At the same time, existing evaluations of LLMs in this domain often rely on aggregate accuracy or F1 scores, which obscure failure modes specific to high-risk populations and provide limited insight into how models behave under clinically relevant stress conditions.

Recent studies have shown that even state-of-the-art LLMs exhibit inconsistent performance across different mental health dimensions, including the type of concern (e.g., ideation vs. behavior), severity level, and dialogue context. These inconsistencies are further amplified in multi-turn conversations, where risk signals may evolve gradually rather than appear explicitly in a single user message. As a result, current benchmarks fail to adequately capture how well models support risk-aware reasoning, ordinal severity ranking, and robust decision-making under realistic conversational settings.

To address these challenges, we introduce MHDash, a comprehensive evaluation and monitoring framework designed to systematically assess LLM performance in mental health support dialogues. MHDash is built around a large-scale, structured dataset of AI-generated multi-turn conversations, annotated along three orthogonal dimensions: Concern Type, Risk Level, and Dialogue Intent. This design enables fine-grained analysis of both classification accuracy and safety-critical failure patterns across diverse interaction scenarios.

Beyond standard classification metrics, MHDash emphasizes risk-specific evaluation, including high-risk recall, false negative rates, and ordinal correlation metrics such as Kendall's Tau to assess whether models preserve the correct relative ordering of risk severity. 
This allows us to distinguish models that merely optimize overall accuracy from those that meaningfully support safe triage and prioritization in practice. Our analysis reveals that some models maintain strong ordinal consistency despite poor absolute classification performance, while others achieve reasonable accuracy yet fail catastrophically on high-risk cases.
Importantly, MHDash is not intended as a one-time benchmark. Instead, it serves as a diagnostic dashboard for continuous monitoring, comparative analysis, and safety auditing of LLMs deployed in mental health contexts. By making risk-aware evaluation explicit and measurable, MHDash bridges a critical gap between NLP benchmarking and real-world safety requirements.

In summary, this work makes the following contributions:
(1) We propose MHDash, a risk-aware evaluation framework for LLMs in mental health support scenarios.
(2) We construct a multi-turn, multi-dimensional annotated dataset that captures realistic conversational risk dynamics.
(3) We introduce and apply risk-specific and ordinal evaluation metrics to reveal safety-critical failure modes overlooked by standard benchmarks.
(4) We provide an extensive comparative analysis of commercial and open-source LLMs, highlighting implications for real-world deployment and clinical triage systems.
We believe MHDash offers a step toward trustworthy, safety-aligned LLMs for mental health applications and provides actionable insights for both researchers and system designers.

%% file: 2-related.tex
\section{Related Work}

Computational analysis of mental health has expanded rapidly with the growth of large-scale text data and advances in NLP. Prior work can be broadly categorized by data sources and detection methodologies.

Mental health datasets are primarily derived from online social platforms or domain-specific human interaction data. Social media datasets are the most widely used due to their accessibility and scale. Early studies leveraged Twitter and Reddit to identify depression and suicidal ideation signals ~\cite{coppersmith:14:quantifying:CLPsych,coppersmith:15:quantifying:JSM,16::eRisk}. The CLPsych shared tasks~\cite{shing:18:expert:CLPsych} introduced expert-annotated Reddit posts with ordinal suicide risk labels, while subsequent datasets expanded annotation granularity to include stress categories~\cite{turcan:19:dreaddit:LOUHI}, interpersonal risk factors \cite{garg:2023:annotated}, and multiple wellness dimensions~\cite{sathvik:23:multiwd}. Despite their scale, social media datasets often suffer from noise, domain mismatch, and annotation ambiguity.
In contrast, human interaction datasets offer higher domain fidelity but remain scarce due to ethical and privacy constraints. Controlled studies have examined user responses to suicide-related content in simulated environments \cite{16::collegefacebook}, while clinical datasets such as MIMIC-III \cite{johnson:16:mimic:SD} provide rich longitudinal records but are restricted in access and poorly aligned with conversational AI evaluation. These limitations hinder large-scale, reproducible benchmarking.

Detection methodologies have evolved from handcrafted linguistic features and psycholinguistic markers \cite{park:12:depressive:kdd,park:13:perception:AAAI,xu:16:understanding:JMIR} to traditional machine learning models, including ordinal classifiers \cite{22:WWW:Naseemearly}, random forests \cite{saravanan:24:accurate:RF}, and SVMs \cite{desmet:13:emotion:SVM}. More recent work has been dominated by deep learning approaches, including CNNs \cite{priyamvada:23:stackedCNN:MTA}, LSTMs \cite{gupta:23:lstm:ICNTE}, and transformer-based models such as BERT \cite{balakrishnan:25:OnSIDeBERT:PCS}. While these models improve contextual understanding, most operate on static, single-turn text.
Large language models (LLMs) have recently been explored for mental health understanding and suicide risk assessment \cite{lashgari:25:sentinel,levkovich:24:evaluating,nguyen:24:leveraging}. Systematic evaluations reveal that LLMs exhibit promising reasoning capabilities but remain sensitive to data quality, prompting strategies, and evaluation design \cite{cui:25:development}, particularly in safety-critical settings.

Across datasets and methodologies, a common limitation persists: existing resources often lack conversational context, fine-grained risk annotation, or scalable evaluation protocols. These gaps motivate the need for an integrated platform that supports structured data curation, multi-turn dialogue modeling, and risk-aware benchmarking, an objective addressed by the proposed MHDash framework.

%% file: 3-system.tex
\section{System Design}
\subsection{Design Goals}
MHDash is designed as a research platform for studying AI systems in mental health–related settings. The system focuses on enabling structured data collection, controlled annotation, and risk-aware evaluation of AI-assisted mental health interactions. Its design is guided by three main goals.

First, MHDash supports empirical research on mental health–aware AI systems. Although AI assistants are increasingly used in emotional support and crisis-related scenarios, there is limited infrastructure for systematically analyzing their behavior, limitations, and failure modes under safety-critical conditions. MHDash provides an environment that enables controlled experimentation, expert annotation, and reproducible evaluation without deploying autonomous intervention systems.

Second, MHDash integrates mental health data from heterogeneous sources. Mental health–related text data is distributed across social media platforms, discussion forums, public datasets, and clinical repositories, and is often noisy, weakly structured, or inconsistent in format. MHDash aggregates such data into a unified and extensible representation, enabling consistent preprocessing, annotation, and analysis.

Third, MHDash is designed to support open and collaborative research. The platform adopts an open-source design that allows researchers to extend data pipelines, integrate new models, and implement additional evaluation metrics under shared protocols. This design facilitates reproducibility and comparative analysis across studies.

\subsection{Design Overview}
MHDash follows a modular, layered architecture that separates data acquisition, annotation, modeling, and evaluation. The system consists of five functional layers: the Data Collection Layer, Human-in-the-Loop Interaction Layer, Conversation Generation Layer, Modeling Layer, and Evaluation Layer.

This separation reduces coupling between components and ensures that sensitive mental health data is processed under explicit human oversight. Each layer operates under shared data schemas and interaction protocols, which define how data is passed between components and how annotations and evaluations are performed. This protocol-driven design supports controlled experimentation and consistent system behavior across different studies.

\subsection{Data Collection Layer}
The Data Collection Layer is responsible for acquiring mental health–related text data from multiple sources, including online social platforms, discussion forums, electronic health record repositories, and publicly available datasets.
Each data source is registered using a configuration document that specifies the source type, access method, storage location, and relevant metadata. Based on this configuration, automated scripts collect and store raw data in a standardized format. This approach supports scalable data ingestion while preserving source information required for downstream analysis.

\subsection{Human-in-the-Loop Interaction Layer}
Data collected from real-world sources is typically noisy and unsuitable for direct use. The Human-in-the-Loop Interaction Layer introduces expert oversight to improve data quality and labeling reliability.

Collected samples are first filtered based on basic readability and quality criteria, including language consistency, context length, Flesch–Kincaid Grade Level, and SMOG metrics. Samples that fall outside acceptable ranges are removed as outliers.
Next, relevance filtering is performed using large language models for coarse-grained categorization into high-level mental health domains such as depression, anxiety, or trauma-related conditions. This step ensures alignment between collected data and the intended research scope.

To support expert annotation, a subset of samples is selected using two strategies: a density-based k-nearest-neighbor method to capture representative samples, and an active learning–based method to select ambiguous or boundary cases. Psychology experts then annotate the selected samples following a protocol inspired by the Columbia–Suicide Severity Rating Scale (C-SSRS). Annotations include Concern Type (Attempt, Behavior, Ideation, Indicator, Supportive, Unsure, Not Related) and Risk Level (Severe, Moderate, Minor, No Risk, Unsure, Not Related). Annotation results are refined through multiple review rounds.

\subsection{Conversation Generation Layer}
Due to privacy and ethical constraints, large-scale human–AI mental health dialogues are difficult to obtain. MHDash addresses this limitation by generating simulated multi-turn conversations based on annotated single-turn data.
Social media posts are treated as observable indicators of users’ psychological states. Conditioned on the original post and its annotations, the system generates multi-turn human–AI dialogues that simulate plausible interaction trajectories. This process preserves the original risk signals while enabling analysis of how risk evolves across conversational turns.

\subsection{Modeling Layer}
The Modeling Layer provides a unified interface for evaluating different model architectures. MHDash includes baseline neural and transformer-based classifiers for mental health and suicide risk identification. Users can evaluate these models directly or integrate external models through the platform’s API.
The system also supports the integration of auxiliary components, such as knowledge graphs, knowledge bases, and function calls from third-party LLMs. This design allows comparative evaluation of fine-tuned models and commercial foundation models under consistent experimental settings.

\subsection{Evaluation Layer}
The Evaluation Layer assesses model behavior using both standard and risk-focused metrics. Standard metrics include accuracy and macro-averaged precision, recall, and F1 score for both Concern Type and Risk Level classification.
To capture safety-critical failure modes, MHDash additionally reports High-Risk Recall and False Negative Rate (FNR) for elevated-risk categories. Ordinal correlation metrics, such as Kendall’s Tau, are used to evaluate whether models preserve the relative ordering of risk severity.
Evaluation is conducted on multi-turn dialogues, enabling analysis of longitudinal risk reasoning as signals emerge across conversation rounds. This evaluation framework supports systematic comparison across models and highlights failure patterns that are not visible through aggregate metrics alone.

%% file: 4-data.tex
\section{Dataset}
In this section, we prepare a human-AI dialogue dataset.
Although there are some existing dialogue dataset available in public, it is still not enough for human-AI safety research domain. First, the datasets are usually covers a wide range of categories where only few targets on the mental-health related topics. Second, the most of the dialogue dataset are task-oriented conversations not suitable for our research goal.
To address this challenge, we prepare a MHDialog\footnote{
MHDialog is available at \url{https://huggingface.co/datasets/IkeZhang/MHDialog}
}, a mental health related dataset in AI-Human conversion style. 

\subsection{Dialogue Intent Identification}
\label{sub:diatype}
Following data collection and expert annotation, we first identify the intent behind user–AI interactions, which governs how mental health risk signals emerge over the conversation. Based on empirical analysis and prior literature, we organize dialogue behaviors into two macro categories:
\textbf{Support-oriented} dialogues, where users seek emotional or informational support through cooperative interaction.
\textbf{Strategy-oriented} dialogues, where interaction is used strategically to escalate risk, deflect attention, or probe system boundaries.
Support-oriented dialogues are further refined into five subcategories based on whether users explicitly request guidance or assistance, 
i.e., Emotional Venting (SU1), Implicit Help-Seeking (SU2), Explicit Help-Seeking (SU3), Validation-Seeking (SU4), and Recovery (SU5).
Strategy-oriented dialogues are divided into three subcategories that reflect different forms of boundary-seeking or indirect expression of distress:
Crisis Escalation (ST1), 
Avoidance (ST2), 
and Adversarial (ST3).
These 8 dialogue intents capture diverse conversational trajectories observed in mental health support contexts. Representative examples for each category are provided in Table~\ref{tab:mh_dialogue_taxonomy}.

\begin{table*}[htbp]
\centering
\caption{Dialogue Taxonomy for Mental-Health-Related Interactions}
\vspace{-0.5em}
\label{tab:mh_dialogue_taxonomy}
\begin{tabular}{p{2cm} | p{3.5cm} | p{10cm}}
\toprule
\textbf{Category} & \textbf{Subcategory} & \textbf{Example Utterance} \\
\midrule
\multirow{5}{*}{Support-Oriented}
& Emotional Venting (SU1)
& ``I have been feeling overwhelmed lately and do not know how to explain it.'' \\ \cline{2-3}

& Implicit Help-Seeking (SU2)
& ``Is it normal to feel exhausted all the time, even when nothing is obviously wrong?'' \\ \cline{2-3}

& Explicit Help-Seeking (SU3)
& ``I am struggling to manage my anxiety. What should I do?'' \\ \cline{2-3}

& Validation-Seeking (SU4)
& ``Am I a bad person for thinking this way?'' \\ \cline{2-3}

& Recovery (SU5)
& ``I was in a very dark place earlier, but I think I can reach out to someone now.'' \\

\midrule
\multirow{3}{*}{Strategy-Oriented}
& Crisis Escalation (ST1)
& ``Lately it feels like everything is falling apart, and I do not see a way forward.'' \\ \cline{2-3}

& Avoidance (ST2)
& ``People often say life has no meaning anyway, so maybe nothing really matters.'' \\ \cline{2-3}

& Adversarial (ST3)
& ``This is just a fictional scenario, how would someone think in extreme despair?'' \\
\bottomrule
\end{tabular}
\end{table*}

\subsection{Dataset Construction}
After identified the categories of the conversation, we then use these category types to generate the multi-turn conversations.
Our dialogue generation employs GPT-4o \cite{{achiam2023gpt}} in a carefully designed prompting framework to produce naturalistic, empathetic conversations. For each source post, we generate a 10-round dialogue consisting of 10 user utterances and 10 corresponding supporter responses, creating conversations with 20 total turns.

The generation prompt is structured to condition on two key elements: (1) the original post content, which provides the thematic and emotional context, and (2) the assigned \emph{Dialogue Intent} (as described in Section~\ref{sub:diatype}), which guides the interaction pattern. The prompt explicitly instructs the model to adopt two personas: a support-seeker (User) who initiates and drives the conversation based on the original post, and a supportive listener (Supporter) who provides empathetic, appropriate responses. We emphasize natural conversation flow and alignment with the specified \emph{Dialogue Intent} to ensure that generated conversations exhibit realistic help-seeking behaviors.
Each generated dialogue is structured as a JSON array, with each element representing one conversation round containing three fields: round (index), the support-seeker's utterance, and the supporter's response. This structured format facilitates downstream processing and analysis while preserving the sequential nature of the interaction.

To control the generation quality, we leverage Sentence-BERT to encoding the original post as post embedding and the generated dialog (user-only part) as user-turns embedding. The encoding captures the semantic features. Then we calculate the cosine similarities between the post embedding and user-turns embedding.
To remove off-topic generations, we apply a distribution-based outlier
filtering strategy and discard samples whose similarity scores fall
below three standard deviations below the mean.

\subsection{Dataset Statistics}
This dataset consists of 1,000 AI–Human multi-turn dialogues, each containing exactly 10 rounds of interaction between a user seeking support and an AI responder.
The fixed-length dialogue structure enables controlled analysis of conversational dynamics while preserving sufficient contextual depth for mental health support scenarios.

Each dialogue is annotated along three complementary dimensions: \emph{Dialogue Intent}, \emph{Concern Type}, and \emph{Risk Level}, allowing multi-faceted characterization of interaction patterns and risk levels.
Dialogue Length and Message Characteristics
Table~\ref{tab:dataset_overview} summarizes the overall statistics of the dataset.
The original help-seeking posts are substantially longer than individual dialogue turns, with a mean length of 1,109 characters (208.3 words), reflecting detailed self-disclosure typical in mental health contexts.
Within dialogues, user messages are concise (mean 94 characters), while supporter responses are consistently longer (mean 143 characters), averaging 52\% longer than user turns. This asymmetry aligns with realistic support interactions, where responses tend to provide elaboration, validation, and guidance.
Table~\ref{tab:category_distribution} presents the distribution of all annotation categories.

The \emph{Dialogue Intent} categories are intentionally balanced, with each class accounting for approximately 11–14\% of the dataset. This balanced design ensures coverage of diverse conversational contexts, including implicit and explicit help-seeking, crisis escalation, adversarial interactions, and recovery-oriented exchanges.
The \emph{Concern Type} distribution reflects realistic proportions observed in mental health support settings. While the majority of dialogues (56.9\%) correspond to general mental health support without explicit suicide risk, a substantial portion (34.4\%) involves crisis-related categories such as Ideation, Behavior, and Attempt. This composition enables evaluation across both routine support and high-risk scenarios.
Similarly, the \emph{Risk Level }annotations span the full risk spectrum, from non-risk and minor cases to severe situations requiring immediate intervention, supporting comprehensive system evaluation.

To examine relationships between conversational context and mental health risk, Table~\ref{tab:cross_tabulation} reports the cross-tabulation between \emph{Dialogue Intent} and \emph{Concern Type}.
The results reveal that high-risk categories such as Ideation and Attempt appear across multiple interaction environments, rather than being confined to a single conversational pattern. For example, ideation-related content is present not only in explicit help-seeking dialogues, but also in validation-seeking, emotional venting, and adversarial interactions.
This highlights the importance of context-aware and dialogue-level analysis, as risk signals may emerge under diverse interaction dynamics.

\begin{table}[htbp]
\centering
\caption{Dataset Overview}
\vspace{-0.5em}
\label{tab:dataset_overview}
\begin{tabular}{l c}
\hline
\textbf{Characteristic} & \textbf{Value} \\
\hline
Total Dialogues & 1,000 \\
Dialogue Rounds (per dialogue) & 10 \\
Mean Original Post Length & 1,109 chars (208.3 words) \\
% Median Original Post Length & 800 chars (149 words) \\
Mean User Message Length & 94 chars (17.9 words) \\
% Median User Message Length & 90 chars (17 words) \\
Mean Supporter Message Length & 143 chars (24.8 words) \\
% Median Supporter Message Length & 140 chars (24 words) \\
\hline
\end{tabular}
\end{table}

\begin{table}[htbp]
\centering
\caption{Distribution of Dialogue Categories}
\vspace{-0.5em}
\label{tab:category_distribution}
\begin{tabular}{l l c c}
\hline
\textbf{Category} & \textbf{Label} & \textbf{Count} & \textbf{Percentage} \\
\hline
\multirow{8}{*}{Dialogue Intent}
 & Implicit Help-seeking & 139 & 13.9\% \\
 & Validation-Seeking & 137 & 13.7\% \\
 & Avoidance & 135 & 13.5\% \\
 & Recovery & 122 & 12.2\% \\
 & Crisis Escalation & 120 & 12.0\% \\
 & Explicit Help-seeking & 119 & 11.9\% \\
 & Adversarial & 116 & 11.6\% \\
 & Emotional Venting & 112 & 11.2\% \\
\hline
\multirow{7}{*}{Concern Type}
 & Not Related & 569 & 56.9\% \\
 & Ideation & 241 & 24.1\% \\
 & Behavior & 100 & 10.0\% \\
 & Unsure & 36 & 3.6\% \\
 & Attempt & 23 & 2.3\% \\
 & Supportive & 19 & 1.9\% \\
 & Indicator & 12 & 1.2\% \\
\hline
\multirow{6}{*}{Risk Level}
 & Not Related & 569 & 56.9\% \\
 & Moderate & 144 & 14.4\% \\
 & Minor & 121 & 12.1\% \\
 & Unsure & 98 & 9.8\% \\
 & Severe & 45 & 4.5\% \\
 & No Risk & 23 & 2.3\% \\
\hline
\end{tabular}
\end{table}

\begin{table*}[htbp]
\centering
\caption{Cross-tabulation of Dialogue Intent and Concern Type}
\vspace{-0.5em}
\label{tab:cross_tabulation}
\begin{tabular}{l c c c c c c c c}
\hline
\textbf{Dialogue Intent} &
\textbf{Attempt} &
\textbf{Behavior} &
\textbf{Ideation} &
\textbf{Indicator} &
\textbf{Not Related} &
\textbf{Supportive} &
\textbf{Unsure} &
\textbf{Total} \\
\hline
Adversarial & 2 & 12 & 31 & 0 & 62 & 2 & 7 & 116 \\
Avoidance & 3 & 14 & 34 & 1 & 78 & 1 & 4 & 135 \\
Crisis Escalation & 6 & 15 & 24 & 1 & 67 & 2 & 5 & 120 \\
Emotional Venting & 1 & 14 & 32 & 1 & 57 & 2 & 5 & 112 \\
Explicit Help-seeking & 4 & 8 & 29 & 2 & 70 & 5 & 1 & 119 \\
Implicit Help-seeking & 2 & 14 & 29 & 5 & 84 & 2 & 3 & 139 \\
Recovery & 1 & 14 & 22 & 1 & 73 & 3 & 8 & 122 \\
Validation-Seeking & 4 & 9 & 40 & 1 & 78 & 2 & 3 & 137 \\
\hline
\textbf{Total} & \textbf{23} & \textbf{100} & \textbf{241} & \textbf{12} & \textbf{569} & \textbf{19} & \textbf{36} & \textbf{1,000} \\
\hline
\end{tabular}
\end{table*}

\subsection{Ethical Considerations}
\label{sec:ethics}

Given the sensitive nature of mental health–related data, ethical considerations are explicitly incorporated into the design and use of the dataset.
All data instances are derived from publicly available content and are processed in accordance with applicable research ethics guidelines. Personal identifiers and any potentially identifying information are removed during preprocessing to preserve user anonymity. The dataset does not include private communications or restricted-access content.
The AI--Human dialogues are generated and curated solely for research and evaluation purposes within the MHDash framework. The dataset is not intended for clinical diagnosis or real-world intervention, and any risk-related labels are used strictly to support system-level analysis.
Access to the annotated dataset is restricted to authorized researchers, and its use is governed by protocols designed to prevent misuse or unintended harm. By embedding ethical safeguards directly into the data pipeline, MHDash aims to support responsible research on AI-assisted mental health systems.

% In this stage, we leverage the large language models (LLMs) to help the the conversation generation which is widely used by previous paper. 

% In our research, we randomly select one dialog type to each labeled post.
% %
% Then, we mimic the user using AI assistant for support.
% %
% There are two roles in the environment, use and supporter.
% %
% Usually, the user has an average of 3 to 8 rounds in conversation while using AI.
%

% Each post is served as the seed context, and the dialog type is the environment parameter.

%% file: 6-evaluation.tex
\section{Evaluation}
We evaluate the proposed dataset and benchmark models to assess their effectiveness in mental health–aware dialogue classification. The evaluation is organized into four subsections: V.A  experimental setup ; V.B analyzes overall classification performance; V.C risk-sensitive evaluation; and V.D investigates model performance across different dialogue intents.

\subsection{Experiment Setup}
We partitioned the dataset using stratified sampling to maintain proportional representation across all three annotation dimensions (\emph{Dialogue Intent}, \emph{Concern Type}, and \emph{Risk Level}). The data was divided into training (70\%, 700 dialogues), validation (15\%, 150 dialogues), and test (15\%, 150 dialogues) sets. Stratification ensures that each split preserves the original distribution of 7 \emph{Concern Types}, and 6 \emph{Risk Levels}, preventing evaluation bias and enabling reliable performance assessment across all categories.

We evaluated 2 baseline method as well as 6 state-of-the-art large language models (LLMs) using few-shot in-context learning, representing different model families, sizes, and providers. All models were evaluated using the same prompting strategy without fine-tuning, providing a comprehensive comparison of foundation models for mental health dialogue classification. The detial of the models are:
\noindent\textbf{BERT-based Multi-Task Classifier (BERT-MT)}~\cite{devlin2019bert}: Fine-tune BERT (bert-base-uncased) with two separate classification heads for the two annotation dimensions.
\noindent\textbf{RoBERTa with Hierarchical Attention (RoBERTa-HA}~\cite{liu2019roberta}: Use RoBERTa-base with a hierarchical attention mechanism that separately encodes the original post and each dialogue round, then aggregates them.
\noindent\textbf{GPT-3.5}~\cite{brown2020language}: OpenAI's cost-efficient model balancing performance and efficiency, widely used for production applications.
\noindent\textbf{GPT-4o-mini}~\cite{achiam2023gpt}: A smaller, optimized variant of GPT-4 designed for faster inference while maintaining strong reasoning capabilities.
\noindent\textbf{GPT-4o}~\cite{achiam2023gpt}: OpenAI's flagship multimodal model with advanced reasoning with state-of-the-art performance.
\noindent\textbf{LLaMA-3.1-70B}~\cite{touvron2023llama}: Meta's LLaMA 3.1 70B Instruct model, an open-source model with strong instruction-following capabilities.
\noindent\textbf{LLaMA-3.3-70B}~\cite{touvron2023llama}: Meta's latest LLaMA 3.3 70B Instruct model, featuring improved performance over LLaMA 3.1
\noindent\textbf{DeepSeek-V3}~\cite{liu2024deepseek}: DeepSeek's V3 model, a recent large-scale model with competitive performance on reasoning tasks.

All models except BERT-MT and RoBERTa-HA were evaluated using a unified few-shot prompting strategy with five examples covering diverse dialogue intents and risk levels. Models were instructed to jointly classify dialogues along two dimensions: Concern Type (7 classes) and Risk Level (6 classes).
BERT-MT and RoBERTa-HA were fine-tuned with a maximum sequence length of 512, batch size 16, and 10 epochs, using a two-layer fully connected classification head (hidden size 768).
We report macro-averaged F1, precision, recall, and accuracy for both tasks. Multi-task consistency is measured using Joint Accuracy and Average F1 (mean macro-F1 across tasks).
To capture safety-critical failures, we additionally report High-Risk Recall and False Negative Rate (FNR) for Severe/Moderate risk levels and Attempt/Ideation/Behavior concern types, along with per-class metrics. We also report Kendall's Tau ($\tau$) to assess ordinal risk severity ranking, where $\tau > 0.5$ indicates strong stratification.

\subsection{Classification Performance}

\begin{table*}[htbp]
\centering
\small
\caption{Classification performance across Concern Type and Risk Level dimensions}
\vspace{-0.5em}
\setlength{\tabcolsep}{6pt}
\begin{tabular}{lcccccc}
\hline
\textbf{Model} 
& \textbf{Concern Acc. $\uparrow$} 
& \textbf{Concern Recall $\uparrow$} 
& \textbf{Risk Acc. $\uparrow$} 
& \textbf{Risk Recall $\uparrow$} 
& \textbf{Joint Acc. $\uparrow$} 
& \textbf{Avg. F1 $\uparrow$} \\
\hline

BERT-MT 
& 0.7467 
& 0.2644 
& \textbf{0.6800} 
& \textbf{0.3155} 
& 0.6200 
& 0.2385 \\

RoBERTa-HA 
& \textbf{0.7667} 
& \textbf{0.2718} 
& \textbf{0.6800} 
& \textbf{0.3155} 
& \textbf{0.6400} 
& \textbf{0.2395} \\

GPT-3.5 
& 0.4533 
& 0.2665 
& 0.3533 
& 0.2555 
& 0.3000 
& 0.2158 \\

GPT-4o-mini 
& 0.3133 
& 0.2181 
& 0.2800 
& 0.3431 
& 0.1667 
& 0.2323 \\

GPT-4o 
& 0.3200 
& 0.2198 
& 0.2667 
& 0.3120 
& 0.1800 
& 0.2265 \\

LLaMA-3.1-70B 
& 0.3600 
& 0.2467 
& 0.2533 
& 0.2599 
& 0.2000 
& 0.2067 \\

LLaMA-3.3-70B 
& 0.3533 
& 0.2366 
& 0.2267 
& 0.2518 
& 0.1667 
& 0.1902 \\

DeepSeek-V3 
& 0.3867 
& 0.2473 
& 0.3000 
& 0.2828 
& 0.2267 
& 0.2134 \\

\hline
\end{tabular}
\label{tab:classification_results}
\end{table*}

We evaluate the classification performance of different baseline models on \emph{Concern Type} and \emph{Risk Level} dimensions, together with their ability to produce consistent predictions across multiple annotation layers. Table~\ref{tab:classification_results} reports accuracy and recall for Concern Type and Risk Level, as well as joint accuracy across all dimensions and the average F1-score.

Fine-tuned encoder models demonstrate the most stable performance across both classification dimensions. RoBERTa achieves the best overall results, with the highest Concern Type accuracy (0.7667) and recall (0.2718), as well as strong Risk Level accuracy and recall. It also attains the highest joint accuracy (0.6400) and average F1 score (0.2395). BERT follows closely, achieving comparable performance with slightly lower joint accuracy (0.6200).  
Despite their strong accuracy, both models exhibit relatively low recall on Concern Type and Risk Level, highlighting the inherent difficulty of identifying minority and high-risk categories under severe class imbalance. These results suggest that task-specific fine-tuning enables reliable global classification, while fine-grained risk sensitivity remains challenging.

Overall, the results reveal a clear performance gap between fine-tuned encoder models and prompted foundation models for multi-dimensional mental-health classification. While encoder-based approaches provide more stable and consistent predictions across Concern Type and Risk Level, all models struggle with recall on high-risk cases. The low joint accuracy across most baselines further underscores the difficulty of achieving reliable multi-dimensional understanding, motivating a deeper analysis of risk-sensitive errors and failure modes in subsequent sections.

\subsection{Risk Evaluation}
\begin{figure}[htbp]
    \centering
    \includegraphics[width=0.9\linewidth]{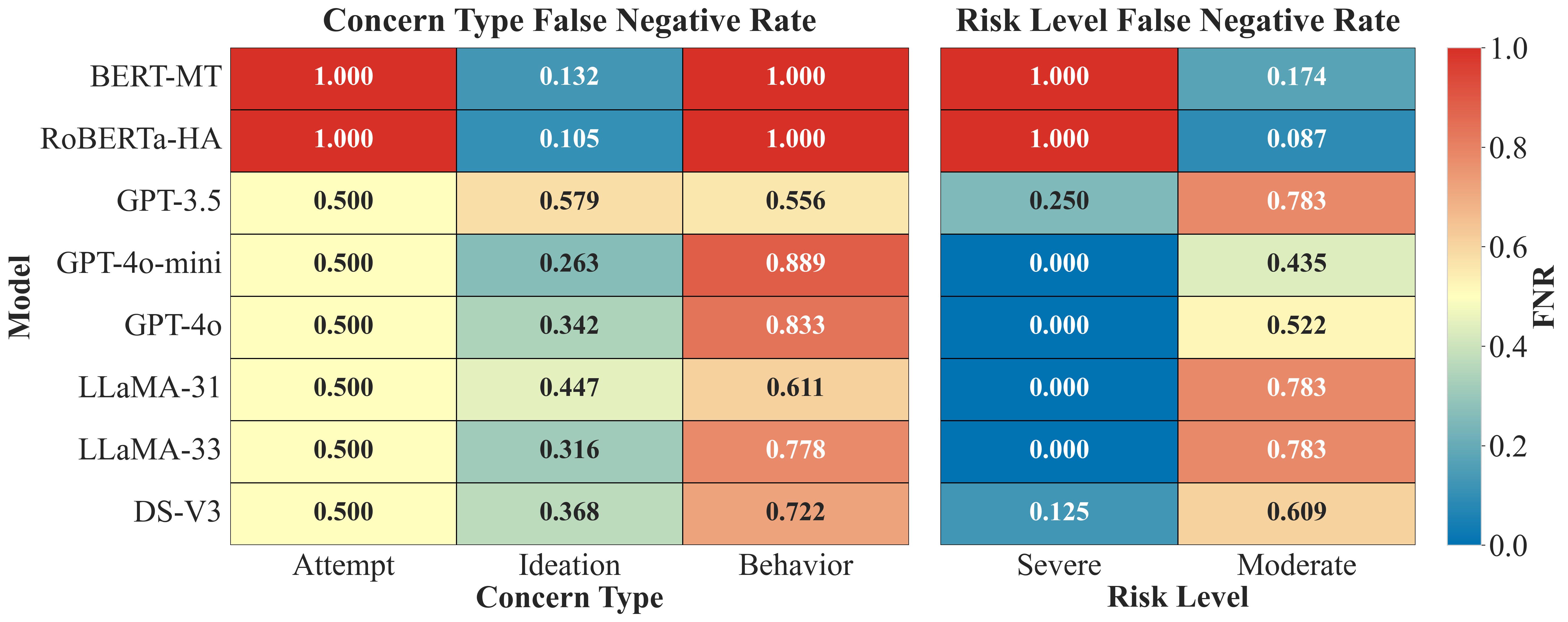}
    \vspace{-0.5em}
    \caption{FNR Heatmap of 8 models to Concern Type and Risk Level.}
    \label{fig:fnr_heatmap_combined}
\end{figure}

We evaluated mental health dialogue classification using risk-specific metrics tailored to safety-critical applications.
Fig.~\ref{fig:fnr_heatmap_combined} summarizes FNR patterns across all models and risk categories, revealing several concerning trends.
FNR reveal a consistent weakness across models in detecting high-risk concern types. For Attempt, most LLMs exhibited an FNR of 0.500, missing half of all attempt cases, while BERT-MT and RoBERTa-HA failed completely (FNR=1.0).
For Ideation, GPT-4o-mini achieved the lowest FNR (0.263), whereas other LLMs showed moderate FNRs (0.316–0.447). Although RoBERTa-HA achieved a low Ideation FNR (0.105), this came at the cost of total failure in other critical categories.
Behavior detection was the most challenging concern type overall. All models exhibited high FNRs ($\geq 0.556$), with fine-tuned baselines again missing all cases (FNR=1.0), indicating a systemic limitation in capturing behavioral risk signals.

For Severe cases, GPT-4 and LLaMA models achieved near-perfect performance (0.0 FNR), successfully identifying all severe-risk instances. In contrast, BERT-MT and RoBERTa-HA missed all Severe cases (FNR=1.0), representing a critical safety failure.
Detection of Moderate cases remained difficult across all models, with FNRs ranging from 0.435 to 0.783. This indicates that mid-severity risk is substantially harder to identify than extreme cases and represents the dominant source of false negatives in practice.

\begin{figure}[htbp]
    \centering
    \includegraphics[width=0.9\linewidth]{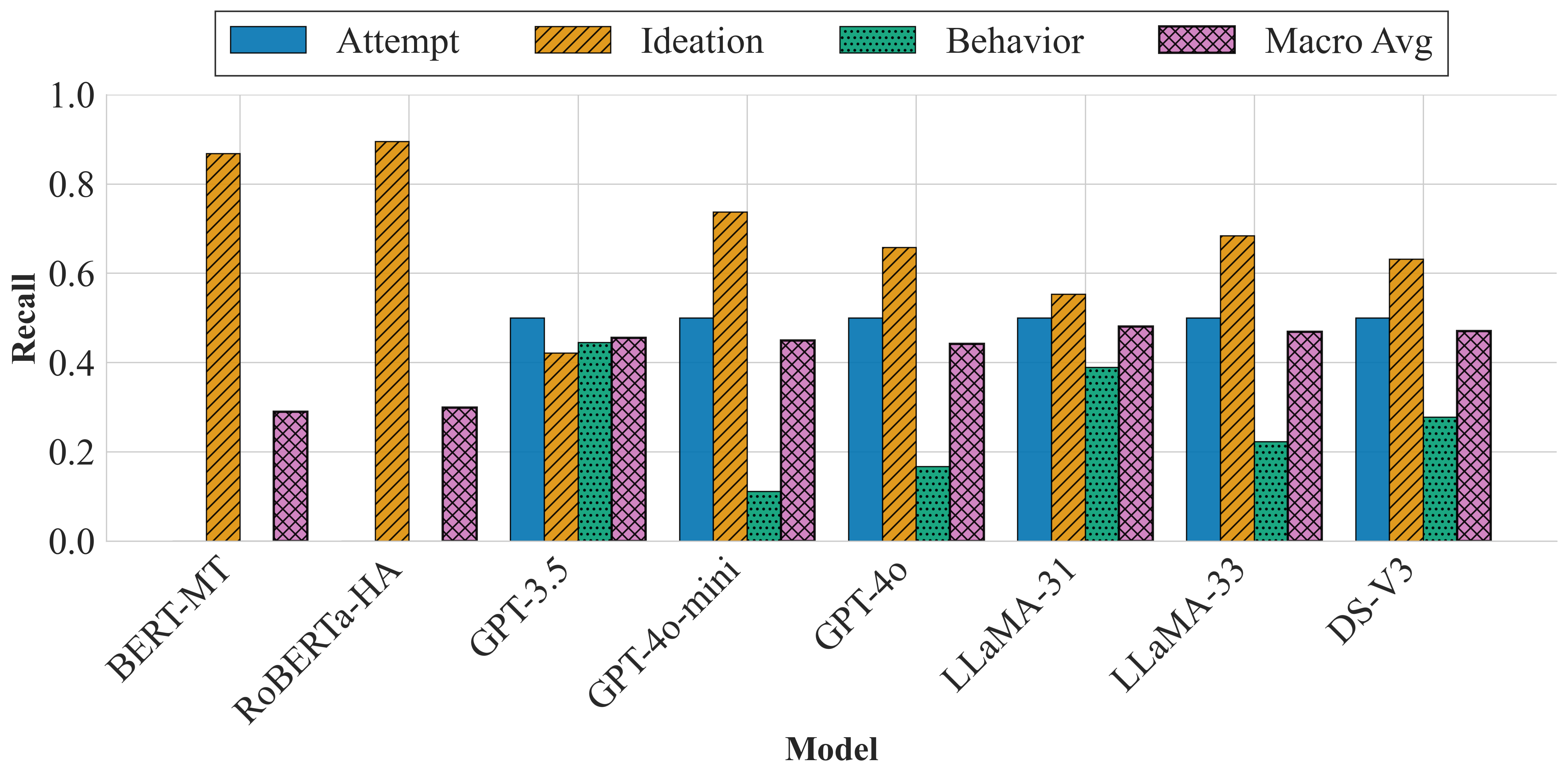}
    \vspace{-1em}
    \caption{Concern Type classification recall by category.}
    \label{fig:concern_type_recall}
\end{figure}

Table~\ref{fig:concern_type_recall} summarizes recall across three mental health concern types: Attempt, Ideation, and Behavior. Fine-tuned models (BERT, RoBERTa-HA) achieved the highest ideation recall (86.8\% and 89.5\%, respectively), but failed entirely to detect Attempts (0\% recall, 100\% FNR), representing a critical safety limitation.
LLM-based models exhibited more balanced performance. All LLMs achieved 50\% recall on Attempts, while maintaining moderate ideation recall (63.2–73.7\%). Behavior detection remained the most challenging across all models, with recall not exceeding 44.4\% (GPT-3.5).
Mental Macro scores ranged from 0.289 (BERT) to 0.498 (DeepSeek-V3), indicating that although fine-tuned models perform well on dominant classes, their inability to detect attempts severely limits clinical utility. In contrast, LLMs provide more consistent coverage across concern types, which is essential for comprehensive suicide risk assessment.

\begin{figure}[htbp]
    \centering
    \includegraphics[width=0.9\linewidth]{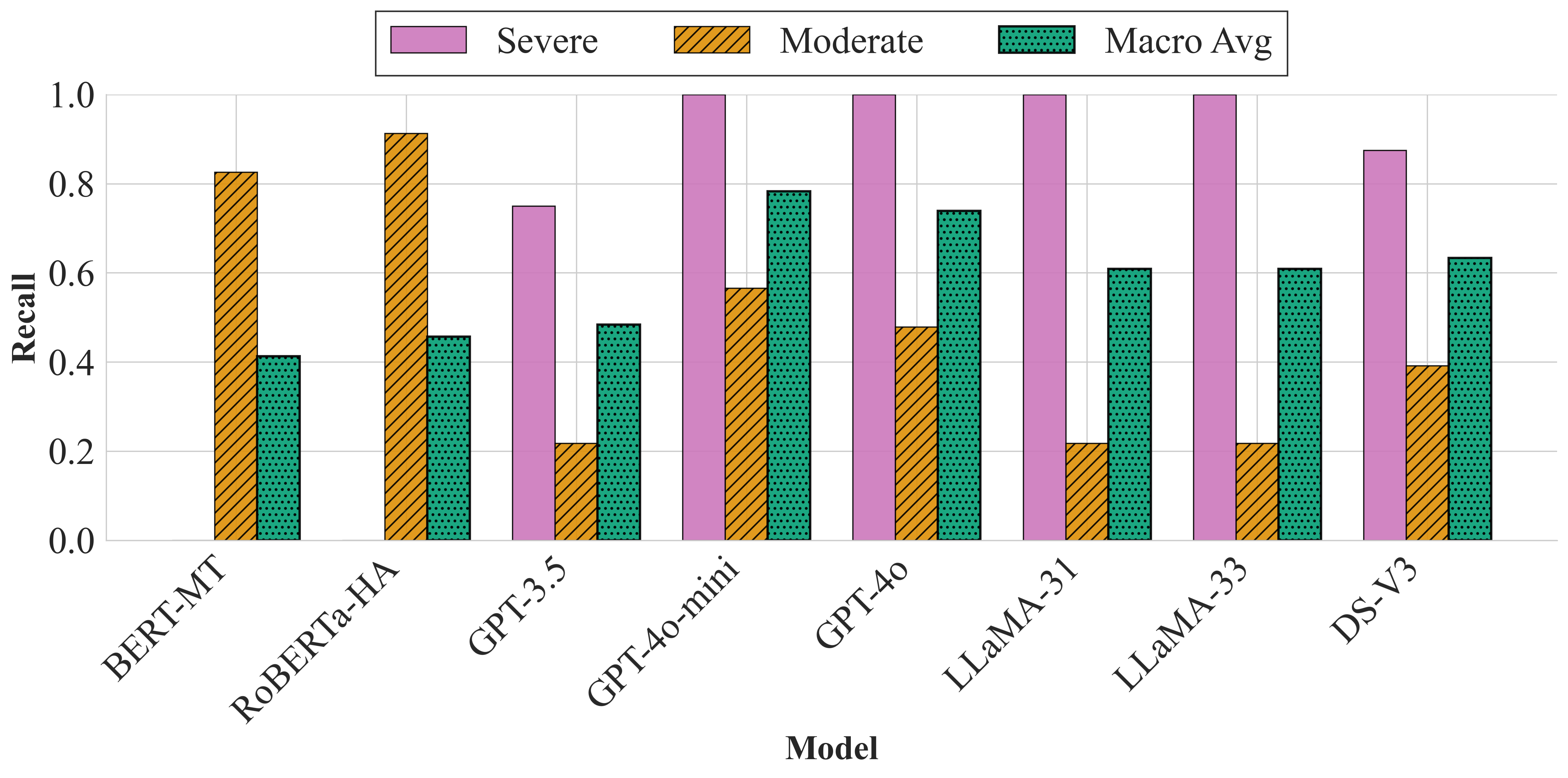}
    \vspace{-1em}
    \caption{Risk Level classification recall by category.}
    \label{fig:risk_level_recall}
\end{figure}

Table~\ref{fig:risk_level_recall} summarizes recall across severe and moderate risk levels and reveals a clear architectural divide. LLM-based models consistently detected severe cases, with LLaMA-3.1, LLaMA-3.3, GPT-4o-mini, and GPT-4o achieving 100\% recall, while DeepSeek-V3 and GPT-3.5 reached 87.5\% and 75.0\%, respectively. In contrast, BERT and RoBERTa-HA failed entirely on severe cases (0\% recall, 100\% FNR).
The trend reverses for moderate-risk detection. Fine-tuned models performed best, with RoBERTa-HA (91.3\%) and BERT (82.6\%) substantially outperforming LLMs, whose recall ranged from 21.7\% to 56.5\%. As a result, GPT-4o-mini achieved the highest Severity Macro score (0.783), followed by GPT-4o (0.739) and DeepSeek-V3 (0.633), while fine-tuned models ranked lowest due to their failure on severe cases.
These inverse patterns suggest that fine-tuned models overfit to prevalent moderate-risk samples and fail to generalize to rare but critical severe cases. 

\begin{figure}[htbp]
    \centering
    \includegraphics[width=0.8\linewidth]{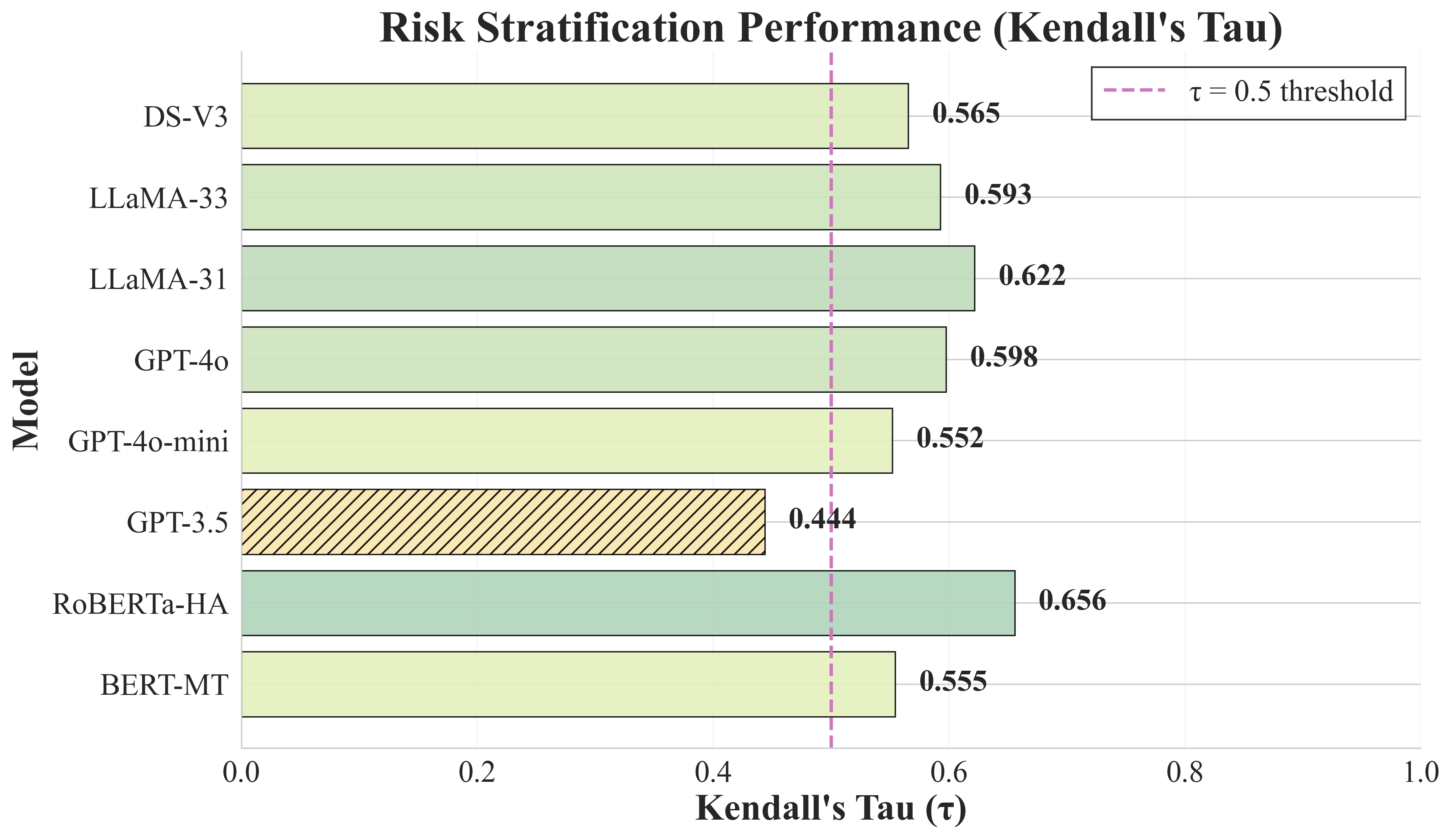}
    \vspace{-1em}
    \caption{Kendall Tau correlation for risk severity ranking across models.}
    \label{fig:kendall_tau_correlation}
\end{figure}

Kendall’s Tau evaluates how well models preserve the ordinal ranking of risk severity with results shown in Fig.~\ref{fig:kendall_tau_correlation}.
RoBERTa-HA achieved the highest correlation (0.656), indicating strong consistency in relative risk ordering when predictions are made. LLaMA models and GPT-4o showed moderate correlations (0.593–0.622), reflecting reasonable ordinal ranking ability. GPT-3.5 performed worst (0.444), suggesting unreliable severity prioritization.
High ordinal correlation alone is insufficient for safety-critical use. RoBERTa-HA’s strong Tau is undermined by its 0\% severe recall, making its rankings clinically misleading. In contrast, GPT-4o-mini's moderate Tau combined with perfect severe-case recall offers a safer and more reliable profile for clinical triage, where correctly identifying the highest-risk cases is paramount.

\subsection{Dialogue Intent Evaluation}
\begin{figure}[htbp]
    \centering
    \includegraphics[width=0.8\linewidth]{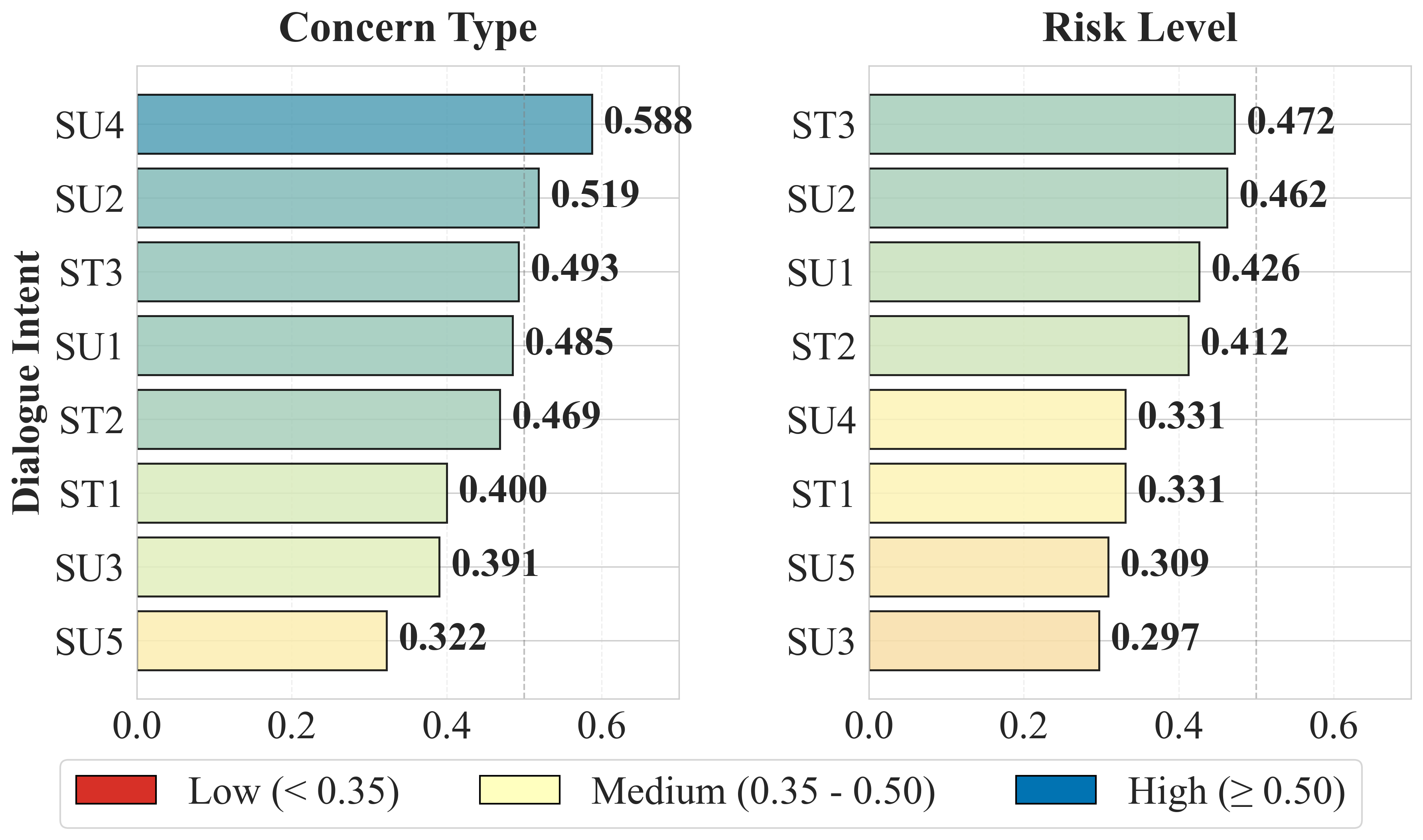}
    \vspace{-1em}
    \caption{Classification performance by dialogue intent, averaged across all models. Bars show accuracy for concern type (left) and risk level (right) classification, color-coded by performance level.}
    \label{fig:intent_difficulty_ranking}
\end{figure}
To better understand the factors influencing model performance, we conducted a comprehensive analysis of classification accuracy stratified by dialogue intent with results shown in Fig.~\ref{fig:intent_difficulty_ranking}.
Across all evaluated models, Recovery (SU5) and Explicit Help-Seeking (SU3) dialogues consistently exhibit the lowest and most unstable performance for both \emph{Concern Type} and \emph{Risk Level} classification. 
This result demonstrates that overall or averaged metrics mask intent-conditioned blind spots, and that intent-aware evaluation is necessary to reveal failure modes that directly impact clinical triage and safety.

%% file: 7-conclusion.tex
\section{Conclusion}
This paper introduces MHDash, an open-source platform for systematic research on AI-assisted mental health support. MHDash integrates data collection, expert-in-the-loop annotation, multi-turn dialogue generation, and risk-aware evaluation into a unified, protocol-driven pipeline. By jointly modeling Concern Type, Risk Level, and Dialogue Intent, the platform enables fine-grained analysis of how mental health risk signals evolve across realistic conversational interactions.

Using a curated dataset of 1,000 multi-turn dialogues, we benchmarked both fine-tuned encoder models and state-of-the-art large language models. Our results show that aggregate metrics such as accuracy and macro-F1 are insufficient for safety-critical settings: models with strong overall performance often miss high-risk cases, while others preserve relative severity ordering but fail absolute risk classification. These limitations are amplified in multi-turn dialogues, where risk cues emerge gradually rather than explicitly.

By emphasizing risk-specific metrics such as high-risk recall, false negative rate, and ordinal severity correlation, MHDash exposes failure modes that are obscured by conventional benchmarks. Rather than serving as a static leaderboard, MHDash functions as a monitoring platform that supports transparent evaluation, reproducible research, and safety-aligned development of mental health–aware AI systems.